\documentclass[10pt,twocolumn,letterpaper]{article}

\usepackage[pagenumbers]{wacv} 

\usepackage{graphicx}
\usepackage{amsmath}
\usepackage{amssymb}
\usepackage{booktabs}

\usepackage[pagebackref,breaklinks,colorlinks]{hyperref}

\usepackage[capitalize]{cleveref}
\crefname{section}{Sec.}{Secs.}
\Crefname{section}{Section}{Sections}
\Crefname{table}{Table}{Tables}
\crefname{table}{Tab.}{Tabs.}

\def\wacvPaperID{2398} 
\def\confName{WACV}
\def\confYear{2025}

\usepackage{multirow}

\newcommand{\dataset}{WASD}

\usepackage{caption}
\usepackage{subcaption}

\begin{document}

\title{ASDnB: Merging Face with Body Cues For Robust Active Speaker Detection}

%
%

\author{Tiago Roxo\\
{\tt\small tiago.roxo@ubi.pt}\\
\and
Joana C. Costa\\
{\tt\small joana.cabral.costa@ubi.pt}\\
\and 
Pedro R. M. In\'{a}cio\\
{\tt\small inacio@di.ubi.pt}\\
\and 
Hugo Proen\c{c}a\\
{\tt\small hugomcp@di.ubi.pt}\\
\and
Instituto de Telecomunicações\\
University of Beira Interior, Portugal\\
}
\maketitle


\begin{abstract}

State-of-the-art Active Speaker Detection (ASD) approaches mainly use audio and facial features as input. However, the main hypothesis in this paper is that body dynamics is also highly correlated to ``speaking'' (and ``listening'') actions and should be particularly useful in \textit{wild} conditions (\textit{e.g.}, surveillance settings), where face cannot be reliably accessed. We propose\break ASDnB, a model that singularly integrates face with body information by merging the inputs at different steps of feature extraction. Our approach splits 3D convolution into 2D and 1D to reduce computation cost without loss of performance, and is trained with adaptive weight feature importance for improved complement of face with body data. Our experiments show that ASDnB achieves state-of-the-art results in the benchmark dataset (AVA-ActiveSpeaker), in the challenging data of WASD, and in cross-domain settings using Columbia. This way,\break ASDnB can perform in multiple settings, which is positively regarded as a strong baseline for robust ASD models (code available at \url{https://github.com/Tiago-Roxo/ASDnB}).

\end{abstract}


\begin{figure}[t]
  \centering
  \includegraphics[width=0.95\linewidth]{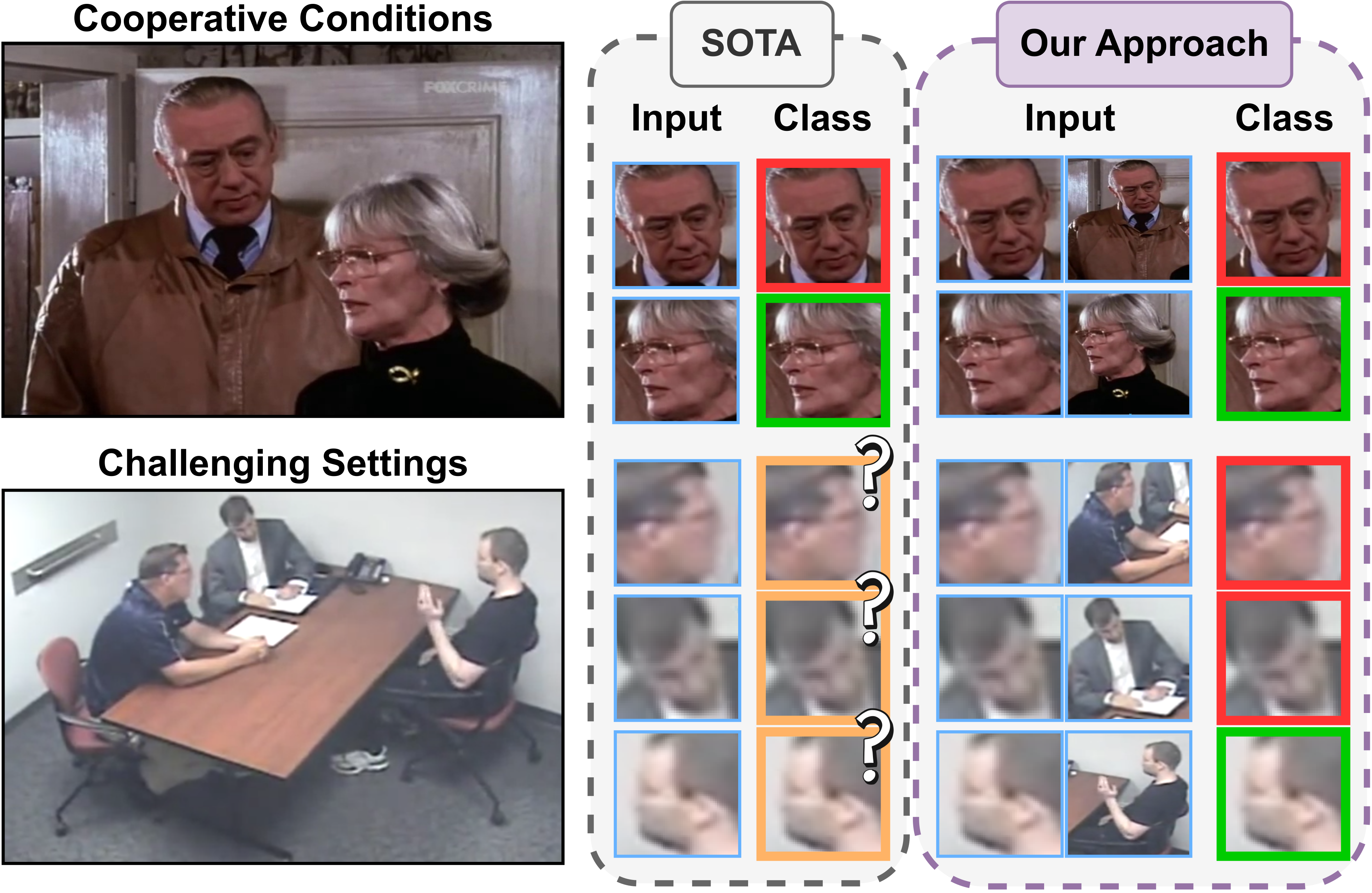}
   \caption{State-of-the-art ASD models solely rely on facial cues as visual input to perform. This approach is only reliable with cooperative (subjects) conditions, increasing uncertainty in model prediction in more challenging settings. Our approach aims to complement face with body cues to create more robust models that are able to perform in cooperative and unconstrained scenarios.}
   \label{fig:init_image}
\end{figure}

\section{Introduction}
\label{sec:intro}

Active Speaker Detection (ASD) aims to identify, from a set of potential candidates, active speakers on a given visual scene~\cite{roth2020ava}, with application in several topics such as speaker diarization~\cite{gebru2017audio,chung2019said, chung2020spot}, human-robot interaction~\cite{kang2023video,skantze2021turn}, automatic video editing~\cite{liao2020occlusion,duan2022flad}, and speaker tracking~\cite{qian2021audio,qian2021multi}. State-of-the-art ASD models typically perform at the video frame level using face data and sound information. Using only facial cues as visual input is a viable strategy due to the correlation of mouth movement and speaking activity, but this approach is only reliable in cooperative and controlled settings. This widely used strategy is motivated by the benchmark ASD dataset AVA-ActiveSpeaker, composed of movies with good audio and face quality. 

Another approach that has not been widely explored for ASD is the use of body information. When people speak or listen they typically use other forms of non-verbal behaviors such as head nodding, hand and body movements, which are not considered by current ASD models. This information increases in importance when face can not be reliably accessed (\textit{e.g.}, face occlusion) or in wilder conditions such as surveillance-like settings~\cite{roxo2024bias}, as shown in Figure~\ref{fig:init_image}. As such, complementing body information with facial cues could improve model robustness to perform in cooperative conditions as well as more challenging settings.

This paper proposes a model that brings Active Speaker Detection and Body (ASDnB) together. In particular,\break ASDnB has the novelty of efficiently combining body with face data using a single visual encoder, merging them at different steps of the extraction, and outputting combined visual features for robust ASD. We modified the visual encoder by splitting the 3D convolution into 2D and 1D to reduce computation cost without loss of performance, and we train ASDnB using adaptive weight feature importance, which results in improved visual encoder extraction and feature complement. Finally, we include temporal modeling in the classifier of ASDnB using bidirectional Gated Recurrent Unit (GRU) layers to maintain the temporal notion for speaker label prediction. Our experiments show that\break ASDnB achieves state-of-the-art results in the benchmark ASD dataset AVA-ActiveSpeaker, as well as in challenging data with degraded audio and face data quality (WASD~\cite{wasd}) and cross-domain settings (Columbia~\cite{chakravarty2016cross}), making it a baseline for robust ASD models. To summarize, the main contributions are:

\begin{itemize}
    \item We announce the first effective combination of body and face data for visual input in ASD, which is a novel approach to create robust models to perform in more challenging settings;
    \item We propose a intra visual encoder combination of dual inputs (face and body) and training with adaptive weight feature importance to effectively combine relevant features for robust ASD;
    \item Ablation studies, experimental evaluation, and performance analysis demonstrate ASDnB state-of-the-art results in the benchmark dataset, AVA-ActiveSpeaker, in the challenging data of WASD, and in cross-domain settings using Columbia.
\end{itemize}

\section{Related Work}
\label{sec:related-work}

\textbf{ASD Context.} Active Speaker Detection is the task to determine the talking speaker from a set of admissible candidates. The benchmark dataset of this area is AVA-ActiveSpeaker~\cite{roth2020ava}, which is based on Hollywood videos totalling almost 38 hours, with demographic diversity and FPS variation, with applications in other areas~\cite{roxo2023exploring}. Several other datasets~\cite{alcazar2021maas, donley2021easycom, kim2021look} were announced since, guaranteeing face access and good audio quality, similar to AVA-ActiveSpeaker, which is not an accurate representation of wild conditions~\cite{roth2020ava}. For ASD in more challenging data, WASD~\cite{wasd} has been recently proposed containing different categories, with varying audio and face quality, ranging from cooperative conditions to surveillance settings. Based on the available data for ASD, current state-of-the-art models heavily rely on face and audio data, combining them using 3D architectures~\cite{chung2019naver}, hybrid 2D-3D models~\cite{zhang2019multi}, and attention mechanisms~\cite{vaswani2017attention, afouras2020self, cheng2020look}. Earlier works are based on a two-step process, where the first focuses on audio with face combination and the second on multi-speaker analysis~\cite{alcazar2020active, kopuklu2021design, zhang2021unicon, alcazar2021maas}, while recently end-to-end models have emerged~\cite{tao2021someone, alcazar2022end, min2022learning, liao2023light, roxo2024bias}. Contrary to existing works, where face is the only visual input for ASD, ASDnB is the first to effectively combine face with body data for robust ASD.

\textbf{Model Enhancement.} Strategies to improve ASD models are typically based on improved feature extraction and combination. Works focus on temporal speech refinement~\cite{alcazar2020active}, inter-speaker and audio-visual relations~\cite{kopuklu2021design, zhang2021unicon}, using Graph Convolutional Networks (GCN)~\cite{welling2016semi} to improve speaker relation representation~\cite{alcazar2021maas, min2022learning, alcazar2022end}, long-term temporal context with audio-visual synchronization~\cite{tao2021someone}, using a reference speech to improve ASD~\cite{jiang2023target}, and changing encoder architectures~\cite{liao2023light}. Despite the different strategies, combining audio with visual features is usually done post encoding using cross-attention approaches~\cite{tao2021someone, jiang2023target} and complemented by temporal modeling. ASDnB is the first ASD model to combine different visual features (face and body), intra encoding, making it more robust to perform in challenging data where face can not be easily accessed.

\textbf{Using Body Information.} Current ASD state-of-the-art models rely on facial cues for visual input given the subject cooperation (face access guaranteed) of the mainstream datasets. However, this is not a viable approach in wilder conditions (such as surveillance settings), where face is not reliably accessed. As such, one potential strategy to improve ASD model robustness is using body information, as explored in other areas. Pedestrian Attribute Recognition (PAR) datasets~\cite{deng2014pedestrian,liu2017hydraplus,li2016richly} are examples of these scenarios, containing person cropped images from surveillance settings, used to identify various attributes under challenging covariates. Works in this area 
focus on different strategies ranging from different architecture combination~\cite{roxo2022yinyang, zhao2018grouping, tang2019improving}, attention-based approaches~\cite{sarafianos2018deep, guo2019visual}, assessing model limitations~\cite{roxo2021wildgender}, and attribute relation importance~\cite{li2015multi, lin2019improving, jia2020rethinking}. In ASDnB, we propose a modification of standard ASD visual encoders, where face and body features are combined at different steps of the extraction, intra encoder, outputting combined visual features.



\begin{figure}[!tb]
\centering
\includegraphics[width=0.99\linewidth]{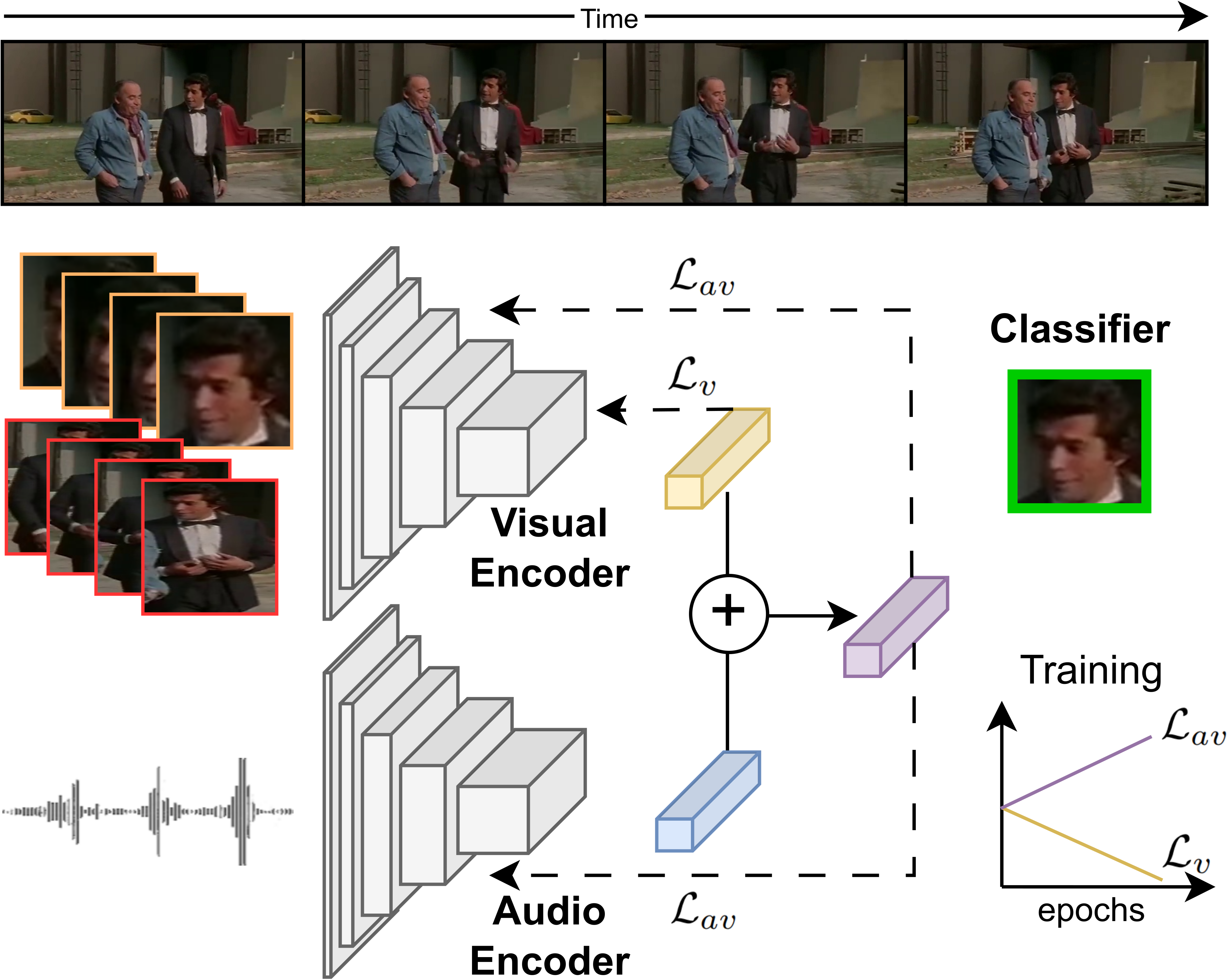}
\caption{
Overview of the ASDnB~architecture. Body and face data are fed into the visual encoder, allowing intra encoder feature fusion and complement, while audio is processed through its respective encoder. Audio and visual features are combined to predict subject speaking label, using an adaptive weighted loss for combined and visual features ($\mathcal{L}_{av}$ and $\mathcal{L}_{v}$, respectively).  
}
\label{fig:main_image}
\end{figure}

\section{ASDnB}
\label{sec:proposed_model}

We propose ASDnB, an model that, for the first time, effectively combines face with body data to perform ASD in cooperative and challenging conditions. We combine dual visual inputs using a single encoder by complementing face and body features at different steps of extraction, and train with adaptive weight to improve feature extraction and combination. The overall architecture of ASDnB is displayed in Figure~\ref{fig:main_image}, with details of each model component in the following subsections.

\subsection{Visual Encoder}

\textbf{Selecting Visual Encoding Approach.} Several strategies for visual encoding in ASD are based on 3D convolutional neural networks, given their effectiveness in extracting spatiotemporal information of face sequences~\cite{kopuklu2021design, alcazar2022end}. These approaches are typically computationally expensive with increased number of parameters, which made other works explore the use of a 3D convolution prior to inputting to a 2D ResNet (typically 18), followed by visual temporal convolution networks~\cite{tao2021someone, jiang2023target}. The key takeaway from the state-of-the-art is that reducing the visual inputs to a 2D context leads to good ASD performance, which can be further extended by splitting the 3D convolution into 2D and 1D to extract the spatial and temporal information, respectively~\cite{liao2023light}. We select this approach for our model since it combines the key strategies of previous ASD works while significantly reducing the number of parameters and computational cost, and without loss of performance.

\begin{figure}[t]
  \centering
  \includegraphics[width=0.99\linewidth]{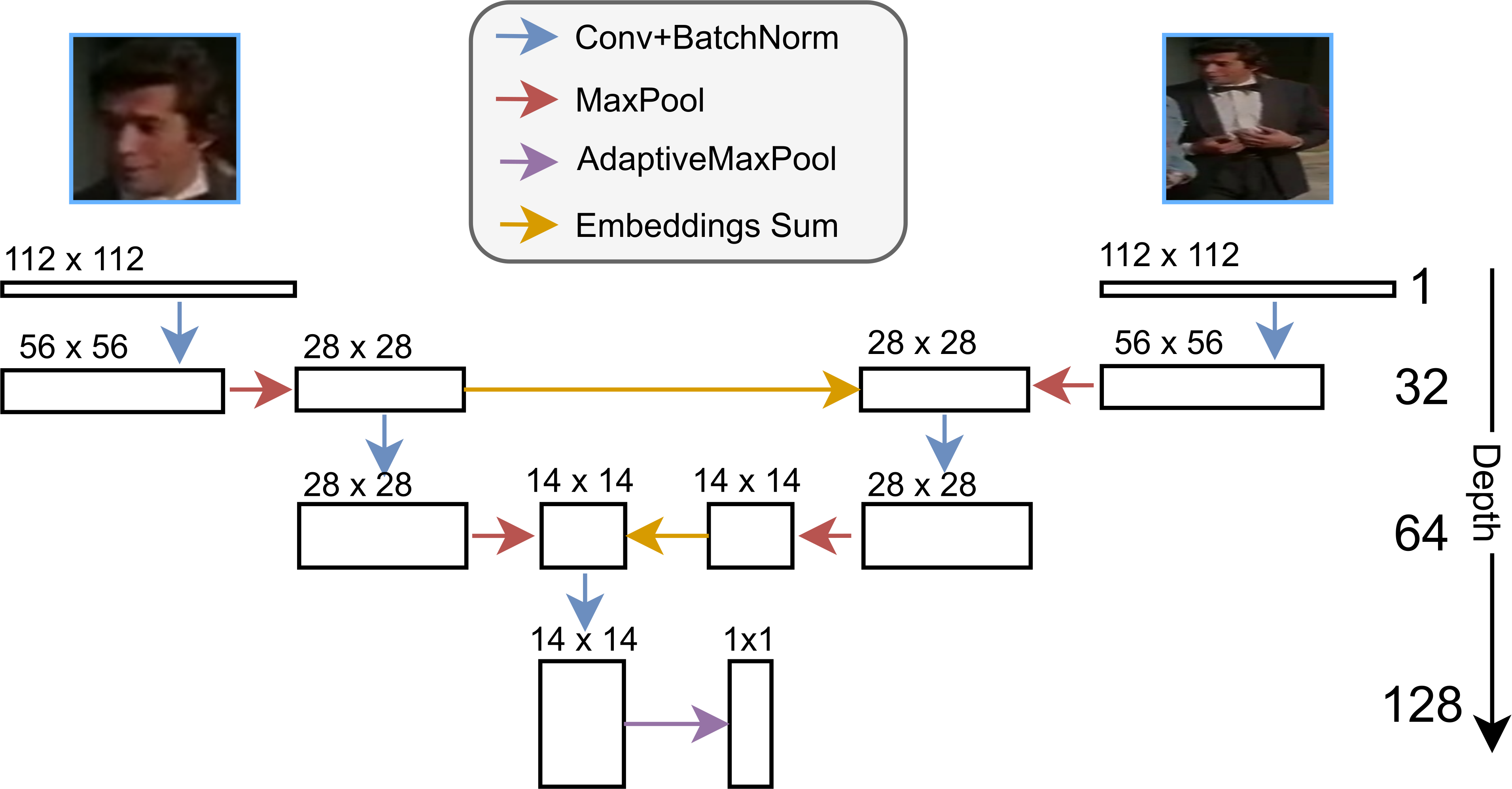}
   \caption{Overview of the flow of face and body combination in ASDnB visual encoder. The first convolution for both inputs downsamples via stride.}
   \label{fig:halfunet_component}
\end{figure}

\textbf{Combining Face and Body.} Contrary to previous works, which only take face as input to visual encoders, we also consider body given its importance to complement facial cues for ASD in more challenging settings (\textit{e.g.}, surveillance conditions). One possible approach is adding two visual encoders to extract face and body features, followed by combining them with audio prior to model classification. However, this does not force the model to consider face and body information in conjunction on feature extraction, resulting in ignoring information complement from the two sources, leading to subpar performances, namely in cooperative settings where face data is reliable~\cite{roxo2024bias}. As such, our motivation is to combine face and body data intra visual encoder to output combined visual feature, using an approach inspired by UNet~\cite{ronneberger2015u} where face and body features are combined at different steps of the extraction, as shown in Figure~\ref{fig:halfunet_component}. We combine face features into body at early extraction steps to make body data as a complement to facial cues since these are the main features for ASD, while recombining body data to face feature at latter stage to reinforce data conjunction. Finally, to ensure spatiotemporal feature extraction abundance at different receptive fields, each convolutional block contains two paths, one with kernel size of 3 and another of 5, given its superiority to other kernel combinations~\cite{liao2023light}.

\subsection{Audio Encoder}

For audio encoding, we adapt the audio signal to serve as input to a 2D encoder by generating Mel-frequency cepstral coefficients (MFCCs), with a sampling rate of 16 kHz, analysis window of 0.025 ms, step between successive windows of 0.010, and a audio representation of 13 cepstrums.  As audio encoder we use ResNet34 with Squeeze-and-Excitation (SE) blocks in its layers (SE-ResNet34), outputting a 128-dimensional audio embedding for subsequent visual and audio conjunction.

\subsection{Temporal Modeling in Classifier}

To improve ASD model performance we apply temporal modeling to the combined multi-modal features from audio and visual encoders. The key motivation is to provide a temporal notion to the model to predict if a subject is talking given that speaking is a continuous action in time, \textit{i.e.} if one subject is talking in a given frame it is more likely to be talking in sequential frames, with a similar logic applying to a non-talking subject. Our approach for ASDnB is shown in Figure~\ref{fig:bigru_detector}, where the combined multi-modal features are obtained by summing visual and audio features, followed by a bidirectional GRU, before passing to a Fully Connected (FC) layer to predict if the candidate is talking.

\subsection{Loss Function}

\textbf{Selecting Loss Strategy.} Existing ASD approaches are typically based on audio and face, with losses tending to rely on the conjunction of both data to assess if a subject is talking. However, recent works~\cite{tao2021someone,liao2023light} explore the inclusion of visual and audio features as sole inputs for losses to complement the cross-modal interaction, leading to improved feature extraction and model performance. In our loss, we also consider visual features importance into it, given that cooperative ASD settings (like AVA-ActiveSpeaker) tend to benefit from assessing facial cues to predict subject talking (\textit{i.e.}, mouth movement heavily relates to talking). 

\textbf{Feature Importance and Combination.} Unlike previous works, we include the notion of gradually increment the importance of combined features (audio with visual) with a relative decrease of visual feature importance throughout training. The key motivation is that, although visual features are important to improve feature combination, its relevance is more crucial in earlier stages of training, with the end goal of ASD models being to assess if a subject is talking via both audio and visual, and not solely based on facial cues. This is particularly important for ASD in more challenging data (\textit{e.g.}, surveillance settings), where the importance of visual features as sole input is not a reliable approach. Furthermore, since our visual encoder has to combine face and body features to output visual embeddings, we want the training to focus early on improving the visual encoder and later at combining audio with visual features. 

Formally, we define ASD as a frame-level classification, where the predicted label sequence is compared with the ground truth via Cross-Entropy:

\begin{figure}[t]
  \centering
  \includegraphics[width=\linewidth]{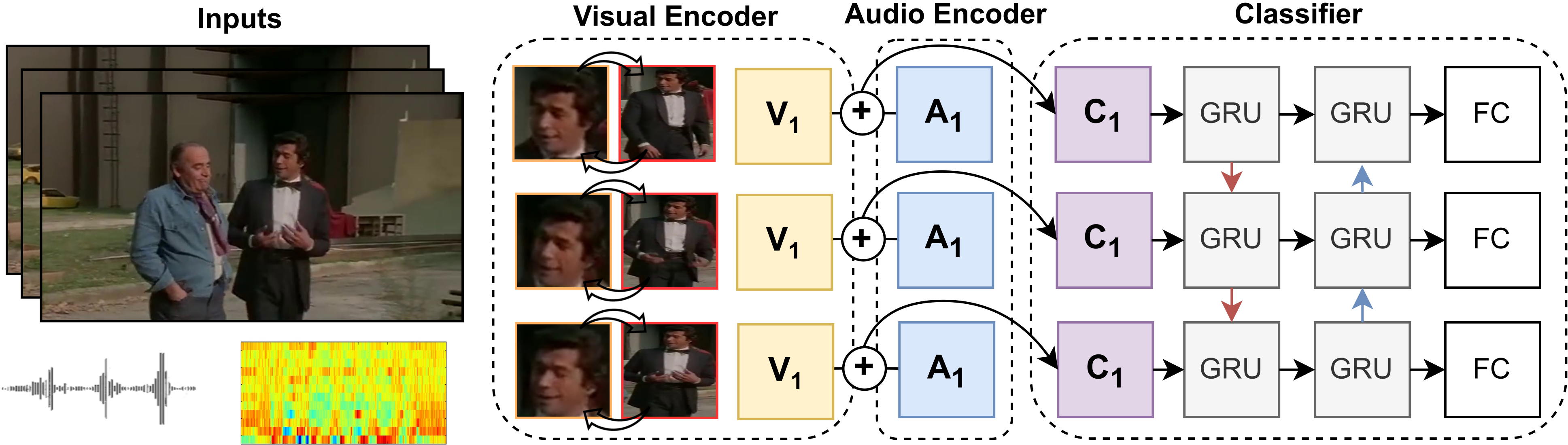}
   \caption{Bidirectional GRUs of visual and audio combination in ASDnB classifier, before inputting to FC layers for speaker classification.}
   \label{fig:bigru_detector}
\end{figure}

\begin{equation}
\mathcal{L} = -\dfrac{1}{T} \sum_{i=1}^{T} (y_{i}~\text{log}(p_{i}) + (1 - y_{i})~\text{log}(1 - p_{i})),
\end{equation}
where $T$ refers to the number of video frames, $p_{i}$ and $y_{i}$ are the predicted and ground truth label for the $i^{th}$ frame, respectively. Finally, the complete loss function is expressed by:

\begin{equation}
\begin{aligned}
\mathcal{L}_{ASDnB} = \alpha~\mathcal{L}_{av} + (1-\alpha)~\mathcal{L}_{v}, 
\end{aligned}
\end{equation}

where $\mathcal{L}_{av}$ and $\mathcal{L}_{v}$ refer to the losses of the combined features (audio with visual) and of the visual feature classification, respectively. $\alpha$ refers to the weight coefficients for combined features, with $\alpha$ starting at 0.5 and incrementing to 1 throughout training, as follows:

\begin{equation}
\begin{aligned}
\alpha = \alpha_{0} + \delta(\varepsilon-1), \\
\end{aligned}
\end{equation}

where $\alpha_{0}$ is set to 0.5 as the initial coefficient importance, $\delta$ is set
to $\dfrac{1}{60}$ as the coefficient decay degree, and $\varepsilon$ refers to the training epoch.

\subsection{Obtaining Body Data}

One of our key contributions relates to combining body with facial cues to retrieve visual features relevant for ASD in varying conditions. However, most ASD datasets do not provide this type of data since current approaches rely solely on face information as visual input. Regarding AVA-ActiveSpeaker, we obtain body bounding box annotations from AVA Actions Dataset~\cite{gu2018ava} (groundwork dataset) and complement them with speaking labels of AVA-ActiveSpeaker, by matching entity id of the original annotations. For Columbia, we use the S3FD face detector~\cite{zhang2017s3fd} based on previous works~\cite{liao2023light,tao2021someone}, resizing its predictions to retrieve the body regions by using twice the width and ending the bottom of the bounding box at three times the predicted height. This approach is only viable for this setting given that subjects in Columbia are cropped to the upper body region (sitting). Regarding WASD~\cite{wasd}, the original dataset already contains body data annotations.

\subsection{Implementation Details}

ASDnB is trained for 30 epochs with an Adam optimizer, with a initial learning rate of 10$^{-4}$, decreasing 5\% for each epoch. All visual data is reshaped into 112 x 112, audio data is represented by 13-dimensional MFCC, and both visual and audio features have an encoding dimension of 128. For visual augmentation, we perform random flip, rotate and crop, while for audio augmentation, we use negative audio sampling~\cite{tao2021someone}. In sum, given a video data during training, a audio track of a new one is randomly selected from the same batch as noise, maintaining the same speaking label of the original soundtrack.

\section{Experiments}
\label{sec:experiments}

\begin{table*}[!tb]
    \centering
    \small
    \renewcommand{\arraystretch}{1.05}
    \caption{Comparison of ASDnB with state-of-the-art models in AVA-ActiveSpeaker.}
    \begin{tabular}{c*{7}{c}}\hline 

      \multirow{2}{*}{\textbf{Model}} &
        \multirow{2}{*}{\textbf{Audio Encoder}} &
      \multirow{2}{*}{\textbf{Visual Encoder}} &
       \multirow{2}{*}{\textbf{Par(M)}} &
       \multirow{2}{*}{\textbf{Pre-training}} &
      \textbf{End-to} &
       \textbf{Body} &
       \multirow{2}{*}{\textbf{mAP}}
       \\

        &  & & &
         & \textbf{End} 
         & \textbf{Data} 
        \\
        \hline\hline
        

        ASC~\cite{alcazar2020active} &  RN18 2D & 
       RN18 2D & 23.3 & $\checkmark$ & $\times$ & $\times$ & 87.1 \\
        
        MAAS~\cite{alcazar2021maas} &  RN18 2D & 
        RN18 2D & 21.7 & $\checkmark$ & $\times$ & $\times$ & 88.8 \\

        UniCon~\cite{zhang2021unicon} &  RN18 2D & 
        RN18 2D & 23.8 & $\checkmark$ & $\times$ & $\times$ & 92.2 \\
        
        TalkNet~\cite{tao2021someone} &  SE-RN34 & 
        RN18+V-TCN & 15.0 & $\times$ & $\checkmark$ & $\times$ & 92.3 \\

        BIAS~\cite{roxo2024bias} & SE-RN34 & 
        RN18+V-TCN & 31.6 & $\times$ & $\checkmark$ & $\checkmark$ & 92.4 \\

        ASD-Trans~\cite{datta2022asd} & RN18 2D & 
        RN18+V-TCN & 15.0 & $\times$ & $\checkmark$ & $\times$ & 93.0 \\

        ASDNet~\cite{kopuklu2021design} & SincDsNet &  
         RNx101 3D & 49.7 & $\checkmark$ & $\times$ & $\times$ & 93.5 \\

        TS-TalkNet~\cite{jiang2023target} & SE-RN34 & 
         RN18+V-TCN & 36.8 & $\times$ & $\checkmark$ & $\times$ & 93.9 \\

         EASEE-50~\cite{easee} & RN50 & 
         RN50 3D & 74.7 & $\checkmark$ & $\checkmark$ & $\times$ & 94.1 \\

         Light-ASD~\cite{liao2023light} & Conv 1D & 
         Conv 2D-1D & 1.0 & $\times$ & $\checkmark$ & $\times$ & 94.1 \\

         SPELL~\cite{min2022learning} &  RN18 2D & 
         RN18+TSM & 22.5 & $\checkmark$ & $\times$ & $\times$ & 94.2 \\

        \hline
        \textbf{ASDnB} & SE-RN34 & 
         Conv 2D-1D & 2.2 & $\checkmark$ & $\checkmark$ & $\checkmark$ & \textbf{94.6} \\
    
        \hline
        
    \end{tabular}
    \label{table:models-performance-dataset-ava}
\end{table*}

\subsection{Datasets and Evaluation Metrics}
\label{dataset}


\textbf{AVA-ActiveSpeaker.} The AVA-ActiveSpeaker dataset~\cite{roth2020ava} is an audio-visual active speaker dataset from Hollywood movies, ranging from 1 to 10 seconds, with 5.3 million face crops, where typically only train and validation sets are used for experiments~\cite{tao2021someone, liao2023light, kopuklu2021design, jiang2023target, roxo2024bias}.


\textbf{WASD.} The WASD~dataset~\cite{wasd} compiles a set of videos from real interactions with varying accessibility of the two components for ASD: \textit{audio} and \textit{face}. With 30 hours of labelled data, \dataset~is divided into 5 categories ranging from optimal conditions to surveillance settings. We report the results on WASD and on each category, following \dataset~experiments.

\textbf{Columbia.} We also consider Columbia~\cite{chakravarty2016cross} following the methodology of Light-ASD~\cite{liao2023light} where models are trained in AVA-ActiveSpeaker, without any additional fine-tuning. Columbia consists of an 87-minute panel discussion video, with five speakers (Bell, Boll, Lieb, Long, and Sick) taking turns speaking, with 2-3 speakers visible at any given time.


\textbf{Evaluation Metrics.} For AVA-ActiveSpeaker and WASD, we use the official ActivityNet evaluation tool~\cite{roth2020ava} that computes mean Average Precision (mAP), while for Columbia we use F1-Score.

\subsection{ASDnB Performance in AVA-ActiveSpeaker}

We compare ASDnB performance with state-of-the-art models in AVA-ActiveSpeaker, in Table~\ref{table:models-performance-dataset-ava}. ASDnB outperforms other models while being \textbf{lightweight and trained end-to-end}. The increased number of parameters from other ASD approaches derive from heavier extraction power of visual inputs and model components (\textit{e.g.}, GCN) to consider author relation which ASDnB simplifies by splitting 3D convolutions into 2D and 1D (lesser computation cost without loss of performance) and using temporal modeling in the classifier. However, the major contribution of ASDnB relative to existing approaches is the \textbf{first efficient combination of face with body data}. The inclusion of body information for ASD is extremely important, in particular for challenging data where face can not be reliably accessed, which is a novel strategy that state-of-the-art models do not yet consider. Only BIAS has previously considered complementing body with face data, but their approach was to treat body has another visual input rather than a complement to face leading to subpar results in AVA-ActiveSpeaker, where face is a reliable visual input for ASD. We are able to efficiently include body for ASD by combining face and body inputs intra encoder, outputting a single combined visual feature that complements facial cues with relevant body movements. Finally, our \textbf{pretraining strategy} is mainly to prepare ASDnB for adequate body information extraction by using WASD~\cite{wasd}, a ASD dataset containing challenging data such as surveillance settings where face is not reliably accessed. Although AVA-ActiveSpeaker is the benchmark dataset, it is not a good representation of \textit{in-the-wild}~\cite{roth2020ava}, with mainly cooperative conditions. As such, for body to have further importance in ASD we prepare the model to perform in harder settings prior to assess ASDnB in more cooperative conditions: AVA-ActiveSpeaker.


\subsection{ASDnB Performance in Other Datasets}

\begin{table}[!tb]
    \centering
    \small
    \renewcommand{\arraystretch}{1.05}
    \caption{Comparison of ASDnB with state-of-the-art models on the different categories of WASD, using the mAP metric. \textit{OC} refers to Optimal Conditions, \textit{SI} to Speech Impairment, \textit{FO} to Face Occlusion, \textit{HVN} to Human Voice Noise, and \textit{SS} to Surveillance Settings. Light refers to Light-ASD, and TS-Talk to TS-TalkNet.}
    \begin{tabular}{cccccc|c}\hline

        \multirow{2}{*}{\makebox[5em]{\textbf{Model}}} &
        \multicolumn{2}{c}{\makebox[5.5em]{\textbf{Easy}}} & 
        \multicolumn{3}{c}{\makebox[5.5em]{\textbf{Hard}}} & 
         \multirow{2}{*}{\textbf{Avg}}\\

        & \textbf{OC} & \textbf{SI} & \textbf{FO} & \textbf{HVN} & \textbf{SS}  \\
        \hline\hline
        
        ASC~\cite{alcazar2020active} & 
        91.2 & 92.3 & 87.1 & 66.8 & 72.2 & 85.7 \\

        MAAS~\cite{alcazar2021maas} & 
        90.7 & 92.6 & 87.0 & 67.0 & 76.5 & 86.4 \\

        ASDNet~\cite{kopuklu2021design} & 
        96.5 & 97.4 & 92.1 & 77.4 & 77.8 & 92.0 \\
        
        TalkNet~\cite{tao2021someone} & 
        95.8 & 97.5 & 93.1 & 81.4 & 77.5 & 92.3 \\

        TS-Talk~\cite{jiang2023target} & 
        96.8 &	97.9 &	94.4 &	84.0 & 79.3 & 93.1 \\

        Light~\cite{liao2023light} & 
        97.8 &	98.3 &	95.4 &	84.7 & 77.9 & 93.7 \\

         BIAS~\cite{roxo2024bias} & 
         97.8 & 98.4 & 95.9 & 85.6 & 82.5 & 94.5 
         \\
    
        \hline
          \textbf{ASDnB} &
          \textbf{98.7} & \textbf{98.9} & \textbf{97.2} & \textbf{89.5} & \textbf{82.7} &  \textbf{95.6} 
          \\

        \hline
        
        \hline
        
    \end{tabular}
    \label{table:models-performance-dataset-wasd}
\end{table}
 
\textbf{Challenging data of WASD.} WASD is divided into categories with incremental challenges to audio and face data, with the most challenging data having face occlusion (FO), audio impairment with background voices (HVN) and surveillance settings (SS), where face access and audio quality is not guaranteed. Table~\ref{table:models-performance-dataset-wasd} shows that ASDnB is superior to all models that only consider face as visual input, in particular for the Hard categories where face and audio quality is impacted, which rarely occurs in AVA-ActiveSpeaker. The biggest performance discrepancy is in surveillance settings (without reliable face input), where only BIAS performance is similar to ASDnB, given its strategy to also consider body data, highlighting the importance of including body information to create robust ASD models. 


\begin{table}[!tb]
    \centering
    \small
    \renewcommand{\arraystretch}{1.05}
    \caption{Comparison of F1-Score (\%) on the Columbia dataset. 
    }
    \begin{tabular}{cccccc|c}\hline

        \multirow{2}{*}{\makebox[5em]{\textbf{Model}}} &
        \multicolumn{6}{c}{\makebox[5.5em]{\textbf{Speaker}}} \\

        & \textbf{Bell} & \textbf{Boll} & \textbf{Lieb} & \textbf{Long} & \textbf{Sick} & \textbf{Avg} \\
        \hline\hline
        
        TalkNet~\cite{tao2021someone} & 
        43.6 & 66.6 & 68.7 & 43.8 & 58.1 & 56.2 \\

        LoCoNet~\cite{wang2023loconet} & 
        54.0 & 49.1 & 80.2 & 80.4 & 76.8 & 68.1 \\

        Light-ASD~\cite{liao2023light} & 
        82.7 & 75.7 & 87.0 & 74.5 & 85.4 & 81.1 \\

        BIAS~\cite{roxo2024bias} & 
        89.3 & 75.4 & 92.1 & 88.8 & 88.6 & 86.8 \\
        \hline


        
        \textbf{ASDnB}&
        \textbf{91.6} & \textbf{81.2} & \textbf{93.1} & \textbf{91.7} & \textbf{90.6} & \textbf{89.6} \\ 
        \hline
        
    \end{tabular}
    \label{table:columbia}
\end{table}

\textbf{Robustness of ASDnB in Columbia.} We also assess the performance of ASDnB in Columbia, following the methodology of Light-ASD~\cite{liao2023light} where models are trained in AVA-ActiveSpeaker, without any additional fine-tuning, and compare with the results reported on Light-ASD, in Table~\ref{table:columbia}. Although Columbia data contains cooperative subjects, the cross-domain evaluation raises challenges for the models. In this context, ASDnB approach to combine body with face and audio information leads to a state-of-the-art performance, and highlights the relevance of complementing face with body data for model robustness to perform in varying conditions such as cross-domain settings.


\subsection{Ablation Studies}

\begin{table}[!tb]
    \centering
    \small
    \renewcommand{\arraystretch}{1.05}
    \caption{Ablation studies on the effect of WASD pretraining, face and body influence towards ASDnB performance (mAP) in AVA-ActiveSpeaker.}
    \begin{tabular}{c*{14}{c}}\hline

        \textbf{Face} & \textbf{Body} & \textbf{Pretrain} &
        \textbf{mAP} \\
        \hline\hline
        
        \multirow{2}{*}{$\times$} &  \multirow{2}{*}{$\checkmark$} &
        $\times$ & 
        83.9 \\ 
        & & $\checkmark$ &  86.5 \\ 
        
        \hline
        \multirow{2}{*}{$\checkmark$} &  \multirow{2}{*}{$\times$} &
        $\times$  & 
        93.7 \\ 
        & & $\checkmark$  & 94.2 \\ 

        \hline
        \multirow{2}{*}{$\checkmark$} &  \multirow{2}{*}{$\checkmark$} &
        $\times$ & 
         94.1 \\
         & & $\checkmark$ & 94.6 \\

        \hline
 
    \end{tabular}
    \label{table:face-body-influence}
\end{table}

\textbf{Feature Influence.} 
Given the ASDnB novelty of body inclusion for ASD, we explore the influence of different features, and pretraining in WASD, in Table~\ref{table:face-body-influence}. The main conclusions are: 1) the variant with only face as visual input and pretraining does not achieves state-of-the-art performances, meaning that body is a necessary feature; 2) pretraining benefits more the ASDnB variant with body than with only face (2.6\% vs 0.5\%), highlighting that pretraining in the challenges of WASD raises more importance to body information relative to facial cues; 3) the combination of body with face information, without pretraining, is the ASDnB variant with best performing results meaning that face with body complement is necessary for ASD but its relevance increases in more challenging data given that the best results require the WASD pretraining, which contains challenges not seen in AVA-ActiveSpeaker. 


\begin{table}[!tb]
    \centering
    \small
    \renewcommand{\arraystretch}{1.05}
    \caption{Variation of audio and visual encoders regarding the number of parameters and model performance in AVA-ActiveSpeaker.}
    \begin{tabular}{c*{14}{c}}\hline

        \textbf{Visual} &
        \textbf{Audio} &
        \multirow{2}{*}{\textbf{Par(M)}} &
        \multirow{2}{*}{\textbf{mAP}} \\

        \textbf{Encoder}  & \textbf{Encoder} \\
        \hline\hline

          RN18+V-TCN &
         Conv 1D & 
          16.5 & 
          92.5 \\ 

         RN18+V-TCN &
         SE-ResNet34 & 
          17.6 & 
          92.7 \\ 

          Conv 2D-1D &
         Conv 1D & 
          1.1 & 
          94.1 \\

            Conv 2D-1D &
            SE-ResNet34 & 
            2.2 & 
            94.6 \\ 

        \hline
 
    \end{tabular}
    \label{table:ablation-encoder}
\end{table}

\textbf{Feature Extraction.} 
We explored variations of audio and visual encoders for ASDnB  and summarized the results in Table~\ref{table:ablation-encoder}. Regarding visual encoders, the approach of splitting 3D convolution into 2D and 1D to extract the spatial and temporal information, respectively, significantly outperforms the standard approach of using a ResNet with temporal convolutional network, while also having lower number of parameters. For ASD visual inputs, facial and body movements are the most relevant aspects, meaning that simpler models capture these notions better, without dispersion to other visual features. For audio encoder, the most robust approach of SE-ResNet34 leads to improved results when combined with the lightweight visual encoder. This is mainly due to the pretraining on WASD, with varying audio quality, that requires more robust extraction of audio features such as distinguish between relevant audio and background voices to combine with adequate visual information.


\begin{table}[!tb]
    \centering
    \small
    \renewcommand{\arraystretch}{1.05}
    \caption{Loss effect on ASDnB performance (mAP) in AVA-ActiveSpeaker. $I_{v}$ refers to the importance of visual features, while $I_{av}$ refers to the importance of combined audio and visual inputs towards ASD prediction. All approaches have Cross-Entropy has the underlying training loss. $\tau$ refers to the temperature coefficient.}
    \begin{tabular}{c*{14}{c}}\hline
        \textbf{Approach} &
        \textbf{$I_{v}$} &
        \textbf{$I_{av}$} &
        \textbf{Extra} &
        \textbf{mAP} \\
        \hline
        \hline

        
        

        Standard & 0 & 1 & $\times$ & 93.1 \\

        TalkNet & 0.4 & 1 & $\times$ & 94.0 \\
        
        Light-ASD & 0.5 & 1 & $\tau$ & 94.2 \\
        
        Our & [0.5-0] & [0.5-1] & $\times$ & 94.6 \\
        \hline
        
    \end{tabular}
    \label{table:ablation-loss}
\end{table}

\textbf{Loss Function.} 
We compare our loss with approaches from other works in Table~\ref{table:ablation-loss}. Standard losses only consider the combined audio and visual features as relevant for ASD, while recent works~\cite{liao2023light, tao2021someone}, focus on complementing this combined loss with weight importance of individual features. The latter strategies tend to perform better since they motivate ASD models to improve visual feature extraction such that visual cues are a reliable source to predict ASD. Light-ASD further improves this aspect by including a temperature coefficient to control feature importance throughout training epochs. Our approach is based on similar concepts but with two key changes: 1) our starting weight importance for visual and combined features increases the relative importance of visual features to motivate a better conjunction of face with body information in the earlier training stages; and 2) we vary visual and combined features importance through training, such that in later stages visual features lose relevance and combining audio with visual features is the strategy for ASD. Our loss translates into better results relative to existing approaches, highlighting the influence of adaptive weight importance for more reliable ASD.


\begin{table}[!tb]
    \centering
    \small
    \renewcommand{\arraystretch}{1.05}
    \caption{Performance of temporal modeling methods in ASDnB classifier.}
    \begin{tabular}{c*{14}{c}}\hline
        \textbf{Method} &
        \textbf{Par(M)} & 
        \textbf{mAP} \\
        \hline
        \hline
        None & 2.02 & 89.8 \\

        Forward LSTM & 2.15 & 93.7 \\
        
        Forward GRU & 2.12 & 93.8 \\
        
        Bidirectional LSTM & 2.28 & 94.4 \\

        Bidirectional GRU & 2.22 & 94.6 \\
        \hline
        
    \end{tabular}
    \label{table:ablation-bigru}
\end{table}

\textbf{Temporal Modeling for ASD.} 
We assess different temporal modeling approaches for ASDnB classifier, in Table~\ref{table:ablation-bigru}. Given the ASD context, speaker prediction benefits from including a temporal relation between frames as shown by the results of not having temporal modeling. Increasing this relation by bidirectional (\textit{vs.} forward) temporal modeling translates into better results, with GRU outperforming LSTM. LSTM tends to be more reliable for long-term information while the simplified version of GRU makes neighboring frames more informative, which is a better approach for ASD.


\subsection{ASDnB Performance Analysis}

\begin{figure}[htp]

\subfloat[Models performance by the number of faces on each frame]{%
  \includegraphics[width=0.9\columnwidth]{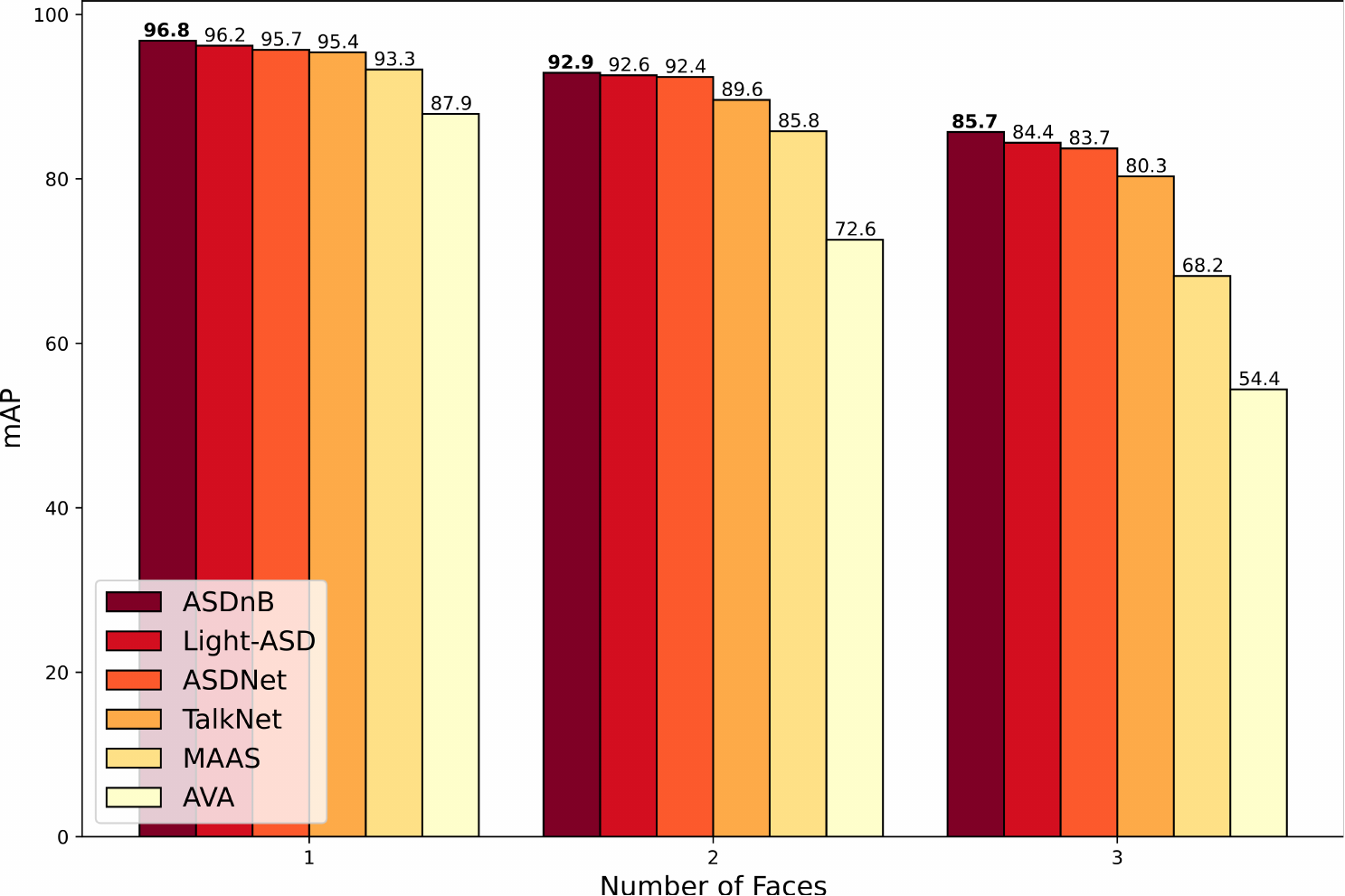}%
}
\vspace{0.4cm}
\subfloat[Models performance by faces size]{%
  \includegraphics[width=0.9\columnwidth]{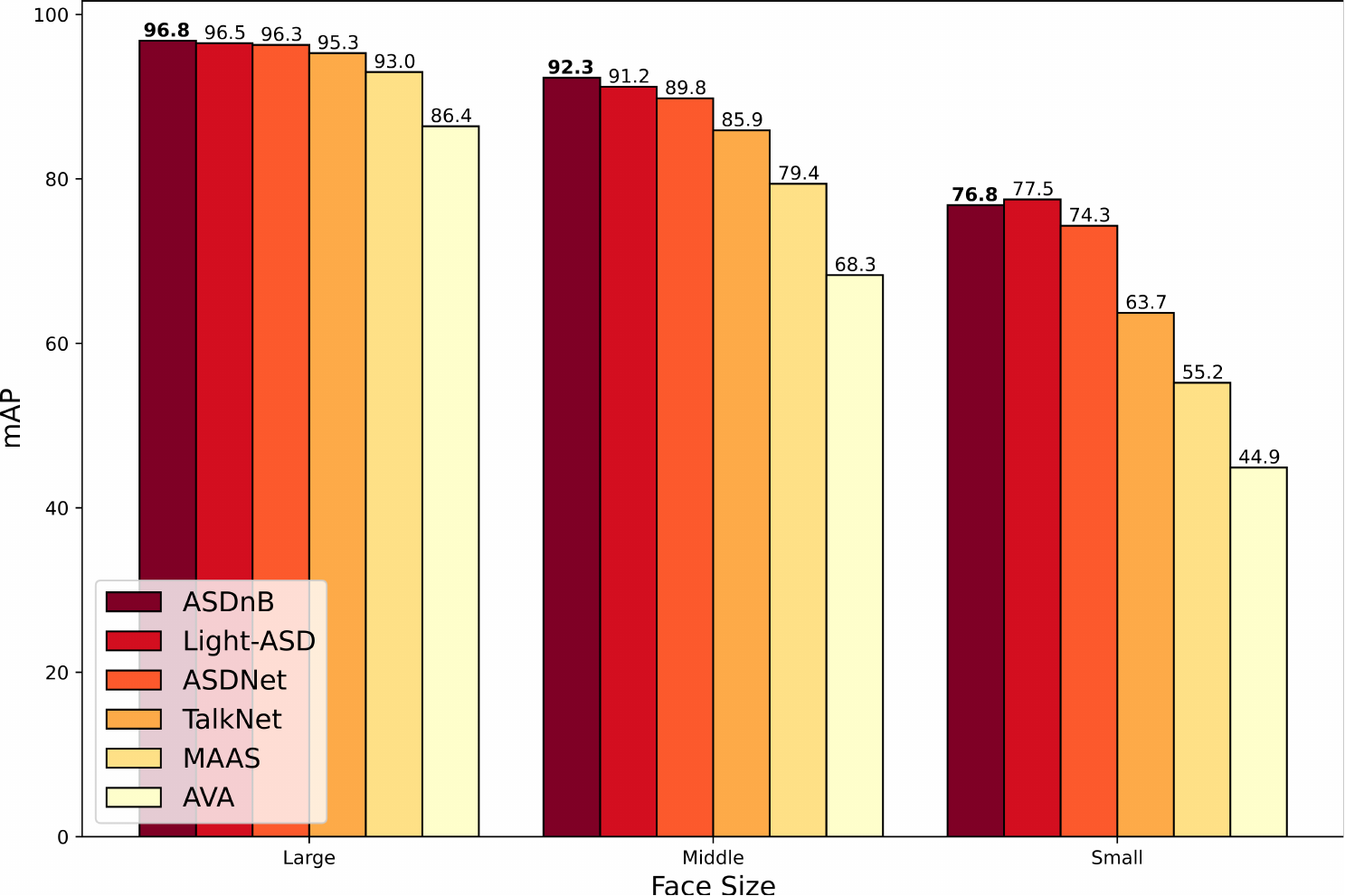}%
}

\caption{Comparison of ASDnB performance relative to ASD state-of-the-art models for (a) number of faces per frame and (b) various face sizes in AVA-ActiveSpeaker.}
\label{face-number-ava}
\end{figure}

\textbf{Face Size and Number of People.} We assess the robustness of ASDnB to deal with variations of AVA-ActiveSpeaker data in Figure~\ref{face-number-ava}, similar to other works~\cite{liao2023light, tao2021someone, alcazar2021maas, kopuklu2021design}. The methodology considers a face as \textit{Small} with width under 64, \textit{Middle} with width between 64 and 128, and \textit{Large} with width over 128, while for the number of people in the scene the data is divided into three mutually exclusive groups (1, 2, and 3) based on the number of faces detected in a frame, totalling 90\% of AVA-ActiveSpeaker data. For all variations ASDnB performance is superior to existing state-of-the-art models, with only a slight underperformance in the smaller face size relative to Light-ASD (76.8\% \textit{vs.}~77.5\%). With smaller faces and no relevant body information ASDnB is not as robust, meaning that there is room for improvement in these settings, namely in background people of a scene. Aside this scenario, ASDnB is an all-around model for ASD, outperforming existing approaches in varying conditions.


\begin{figure*}[!tb]
\centering
\includegraphics[width=0.95\textwidth]{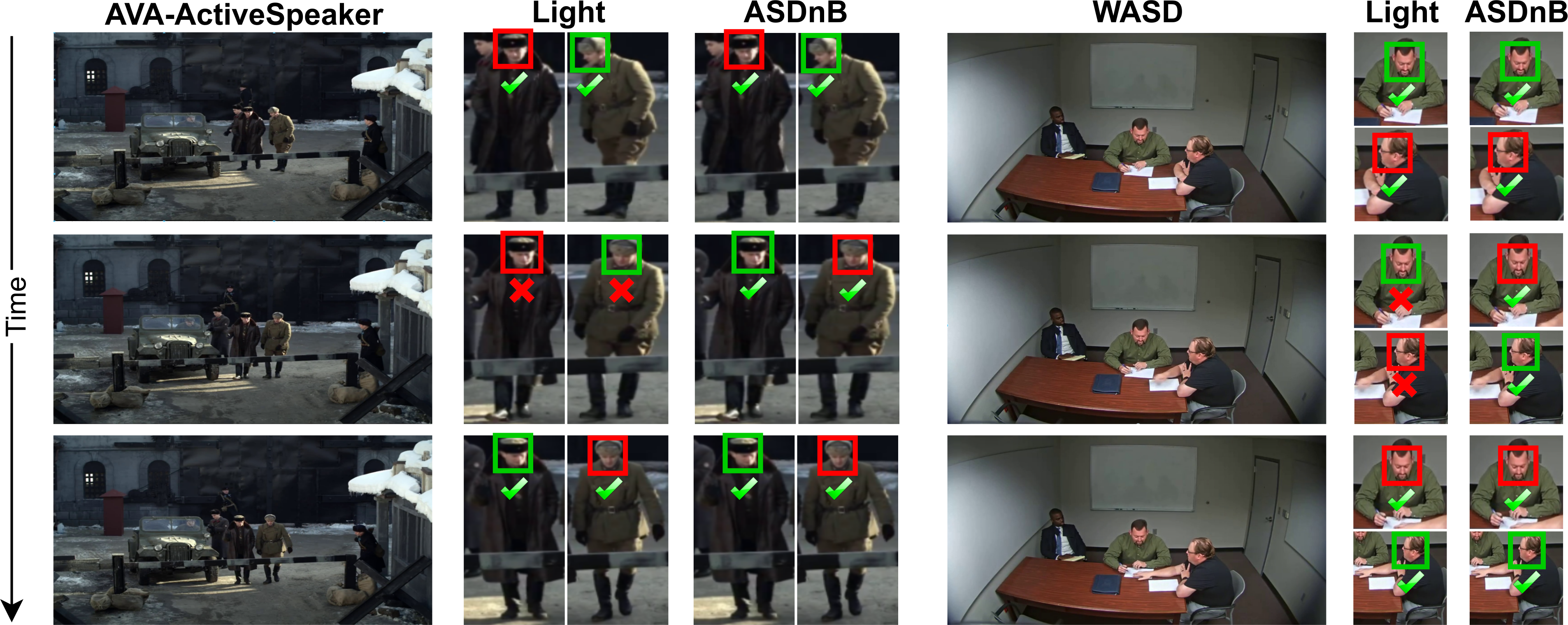}
\caption{
ASDnB and Light-ASD (Light) qualitative performance assessment in challenging scenarios of AVA-ActiveSpeaker and WASD. Red bounding boxes denote model prediction of subject not talking, green to speaking, and predictions with red cross denote missclassification relative to the ground-truth. In both examples with subjects far from the camera, Light-ASD misclassified the switch of speakers while ASDnB was more resilient by analysing the hand movement that preceded subject speaking.
}
\label{fig:light_halfunet_qualitative}
\end{figure*}

\begin{figure}[!tb]
    \centering
    \includegraphics[width=0.9\columnwidth]{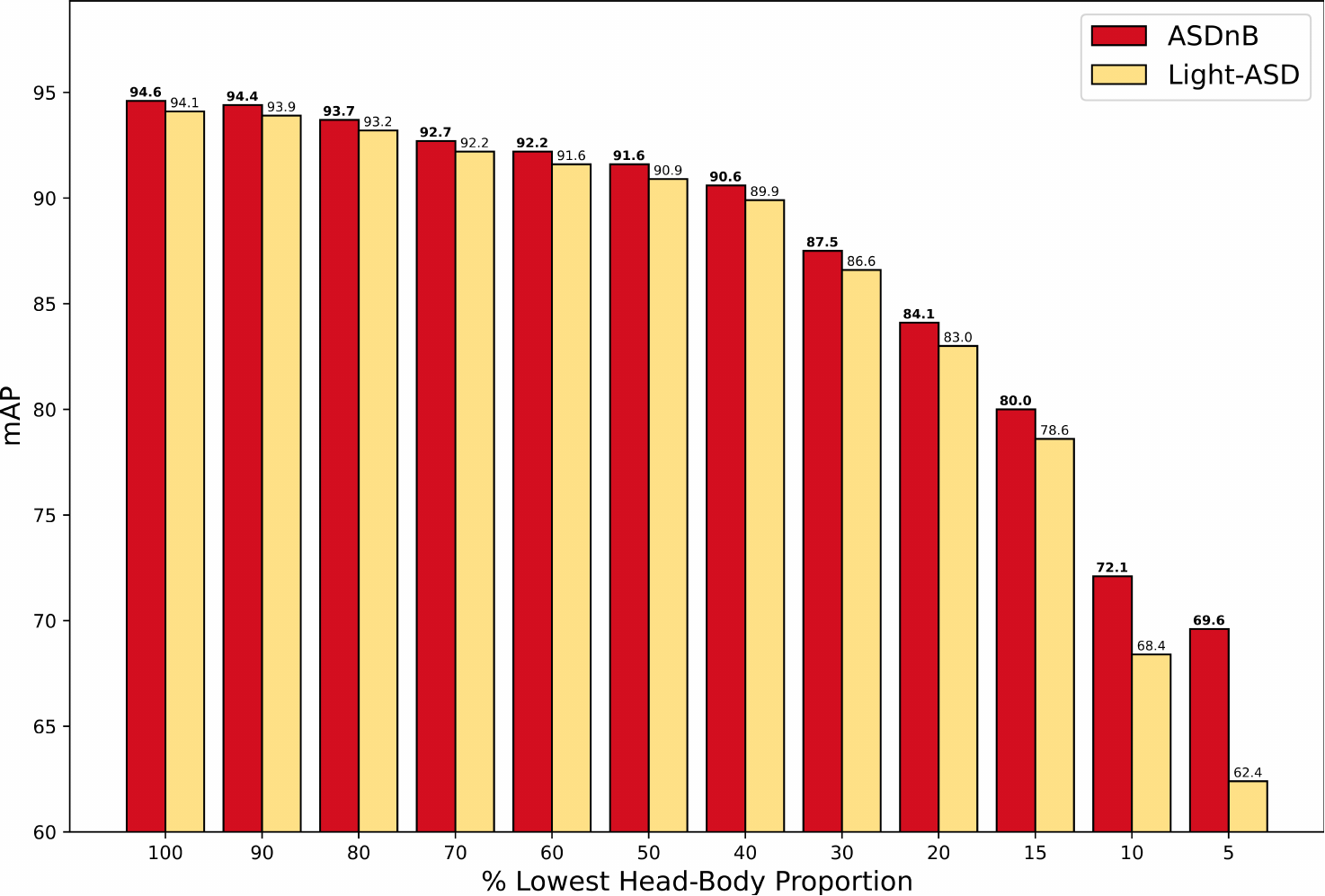}
    \caption{Relative body importance of ASDnB and Light-ASD with decremental head-body area proportion in AVA-ActiveSpeaker.}
    \label{fig:HBP_AVA_bars}
\end{figure}

\textbf{Relative Body Importance.} To further explore the importance of body for ASD, we compare the performance of ASDnB and Light-ASD with varying head-body proportion (HBP), in AVA-ActiveSpeaker, in Figure~\ref{fig:HBP_AVA_bars}. We use all the available testing data in the first pair of bars, and use less data moving left to right on the x-axis, corresponding to lower HBP values. For instance, in the pair of bars at 20\%, we are using the data with the lowest 20\% HBP meaning that these are settings where the face is significantly smaller than body. The results show that, with a decrease of HBP, ASDnB performance is progressively better relative to Light-ASD, highlighting that the absence of reliable face access raises importance in body information. This is particularly predominant in wild conditions such as surveillance settings, meaning that ASDnB strategy is a viable approach to increase ASD robustness to perform in such conditions. Although ASDnB performs worse in smaller faces than Light-ASD (Figure~\ref{face-number-ava}), this aspect is mitigated with relevant body information meaning that the underperformance of ASDnB in such conditions is mainly due to background people without visible body. With relevant body information ASDnB is able to outperform Light-ASD, which reinforces the notion that body data is a relevant feature to use in challenging ASD data (WASD~\cite{wasd}), or when the subject is not close to the camera (instances of AVA-ActiveSpeaker).

\textbf{Qualitative Analysis.} We complement our experiments with a qualitative analysis of ASDnB and Light-ASD in challenging scenarios of AVA-ActiveSpeaker and WASD, in Figure~\ref{fig:light_halfunet_qualitative}. The considered scenarios contain subjects far from the camera and in suboptimal cooperative settings (top-down view) which makes it harder to predict who is speaking using only facial cues. In both examples, Light-ASD misclassified the switch of speakers while ASDnB was more resilient by analysing the hand movement that preceded subject speaking. The results support the importance of body analysis for ASD in wild conditions, where face can not be reliably accessed, making ASDnB a viable baseline for robust ASD models.

\section{Conclusion}
\label{sec:conclusion}

This paper describes ASDnB, a lightweight multi-modal model that, for the first time, efficiently combines face with body information for Active Speaker Detection. The key contribution of our proposal relates to combining face and body features at different feature extraction steps, inspired by the UNet approach, yielding state-of-the-art performance both on cooperative conditions (benchmark dataset AVA-ActiveSpeaker) and on more challenging settings (WASD and cross-domain of Columbia). The obtained results show that complementing body information with facial cues is of utmost importance for ASD robustness, and is particularly important for \emph{wild} conditions (\textit{i.e.}, surveillance settings), where state-of-the-art models do not reliably perform.

\section*{Acknowledgments}

This work was supported by the Project UIDB/50008/2020, FCT Doctoral Grants 2020.09847.BD and 2021.04905.BD, and Project CENTRO-01-0145-FEDER-000019.



\end{document}